# InfyNLP at SMM4H Task 2: Stacked Ensemble of Shallow Convolutional Neural Networks for Identifying Personal Medication Intake from Twitter

**Jasper Friedrichs, MS[1], Debanjan Mahata, PhD[1], Shubham Gupta, BS[2]**
[1]Infosys Limited, Palo Alto, CA; [2]Indian Institute of Technology Guwahati, Assam, India

**Abstract**

*This paper describes Infosys's participation in the "2nd Social Media Mining for Health Applications Shared Task at AMIA, 2017, Task 2". Mining social media messages for health and drug related information has received significant interest in pharmacovigilance research. This task targets at developing automated classification models for identifying tweets containing descriptions of personal intake of medicines. Towards this objective we train a stacked ensemble of shallow convolutional neural network (CNN) models on an annotated dataset provided by the organizers. We use random search for tuning the hyper-parameters of the CNN and submit an ensemble of best models for the prediction task. Our system secured first place among 9 teams, with a micro-averaged F-score of 0.693.*

**Introduction and Background**

Social media has become an ubiquitous source of information for a variety of topics. Right from information related to daily events, personal rants, to expressions of intake of medicine and adverse drug reactions are readily available in publicly accessible social media channels and forums such as Twitter[1], DailyStrength[2], MedHelp[3], among others. Huge amounts of data made available on these platforms has become a useful resource for conducting public health monitoring and surveillance, commonly known as pharmacovigilance[1]. The work presented in this paper aims at identifying intake of personal medication expressed by a user in Twitter. The broader perspective of such a system is to aid in developing automated methods for performing pharmacovigilance activities in social media in order to study the effects of medicine on an individual and specific cohorts[2].

The key to the process of identifying tweets mentioning personal intake of medicine and to draw insights from them is to build accurate text classification systems. The effectiveness of developing classifiers has already been shown to be useful in identifying adverse drug reactions expressed in Twitter[3]. However, mining social media posts comes with unique challenges. Microblogging websites like Twitter pose challenges for automated information mining tools and techniques due to their brevity, noisiness, idiosyncratic language, unusual structure and ambiguous representation of discourse. Information extraction tasks using state-of-the-art natural language processing techniques, often give poor results for tweets. Abundance of link farms, unwanted promotional posts, and nepotistic relationships between content creates additional challenges.

The main objective of the task presented in this paper is to categorize short colloquial tweets into one of the three categories, (a) personal medication intake - tweets in which the user clearly expresses a personal medication intake/consumption, (b) possible medication intake - tweets that are ambiguous but suggest that the user may have taken the medication, and (c) non-intake tweets that mention medication names but do not indicate personal intake. Towards this goal, we design and implement a deep learning based classifier - stacked ensemble of shallow convolutional neural networks, trained on the annotated data provided by the task organizer. Our system performed best among 9 teams, with a micro-averaged F-score of 0.693. Next, we give a detailed description of the submitted system and point to the relevant literature associated with them, whenever necessary.

**Methods**

Deep learning systems have recently shown to achieve top results in shared tasks related to natural language processing on tweets4. Historically, ensemble learning has proved to be very effective in most of the machine learning tasks including the famous winning solution of the Netflix Prize[5]. Ensemble models can offer diversity over training data splits, random initialization of the same model or model architectures, and a combination of multiple average or low performing learners to produce a robust and high-performing learning model. A convolutional neural network (CNN)[6] is a deep learning architecture, that has shown strong performance on sentence-level text classification. Even fairly simple CNNs evaluate at a level of or even better than more complex deep learning architectures[7]. Therefore, we designed and implemented a stacked ensemble of shallow convolutional neural networks (Figure 1) for solving the classification task presented in this paper. The main intuition behind developing such an ensemble was to take the best of all worlds. Next, we explain stacked ensemble of CNNs.

---

[1] http://twitter.com
[2] http://www.dailystrength.org
[3] http://www.medhelp.org/

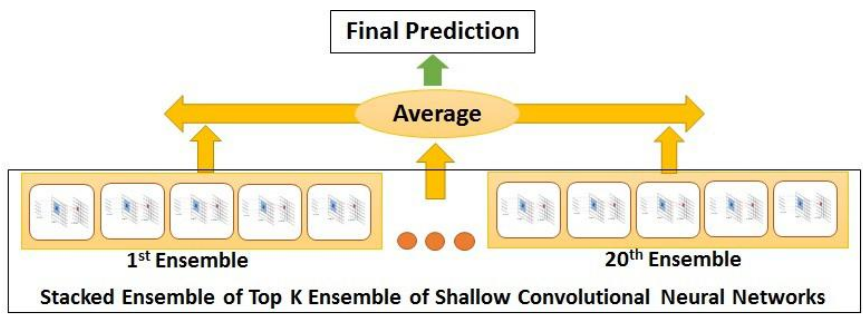

**Figure 1.** A Stacked Ensemble of 100 (20 x 5) shallow convolutional neural networks.

*Stacked Ensemble of Shallow Convolutional Neural Networks*

Figure 1, shows the architecture of the final stacked ensemble of CNNs that we use in predicting the outcome of the task presented in this paper. We train a standard shallow CNN model [6] on each fold while performing 5-fold cross validation on our training set. We take the output prediction of each of these models and average them to create an ensemble of 5 models. We further train twenty such ensembles and take the output of each ensemble and average them to create our stacked ensemble of top 20 ensemble of shallow CNNs. In general, we can take top K such ensembles and create a stacked ensemble of top K ensemble of shallow CNNs.

*Random Search on Hyperparameters*

In order to get the best results from any classification model, hyperparameter tuning is a key step and CNNs are no exception. While the existing literature offers guidance on practical design decisions, identifying the best hyperparameters of a CNN requires experimentation. This requires evaluating trained models on a cross-validation dataset and choosing the best hyperparameters manually that produce best results. Automated hyperparameter searching methods like grid search, random search, and Bayesian optimization methods are also popularly used. In our presented system we use random search[8], to explore the hyperparameters of a shallow CNN architecture and form an ensemble of the best models, which we refer to as a stacked ensemble. Next, we share the detailed output and analysis of our experiments.

*Dataset and Data Preprocessing*

The organizers provided 8000 annotated tweets as a training dataset and 2260 additional tweets as development dataset. We collected the tweets using the script provided along with the dataset, by querying Twitter's API. However, we could not collect all the tweets as some of them were not available at the moment when we executed our collection process. Later, the organizers also shared the test dataset, that was used for calculating the final scores of the submitted models. A distribution of tweets provided for each class and the mapping of each class is shown in Table 1. It is to be noted over here that for training our models, we combine the training and development dataset provided and treat it as our training dataset, therefore learning our models using 9663 tweets with 5-fold cross validation.

|       | Class 1 | Class 2 | Class 3 | Total |
|-------|---------|---------|---------|-------|
| Train | 1847    | 3027    | 4789    | 9663  |
| Test  | 1731    | 2697    | 3085    | 7513  |

**Table 1.** Shared task data distribution. Class 1, 2 and 3 represent *personal medication intake*, *possible medication intake*, and *no medication intake*,

We use Spacy[4] for all our data preprocessing and cleaning activities. We do not remove stopwords. Each document in our training and test dataset is converted to a fixed size document of 47 words/tokens. We use two pre-trained word embeddings - godin[9] and shin[10], shared by the authors. Each of these embeddings are of 400 dimensions. Each word in the input tweet is represented by its corresponding embedding vector, when present in the vocabulary of the model.

*Hyperparameters for the CNNs*

We use Xavier weight initialization scheme[11], for initializing the weights of the CNNs. Adam with two annealing restarts has been shown to work faster and perform better than SGD in other NLP tasks[12]. Therefore, we use the same as our optimization algorithm. We use five filters with varying filter sizes in the convolution layer and use dropout during the training process. The models are implemented using TensorFlow[5]. The entire ranges of the hyperparameters that we give to our random search procedure is shown in Table 2. The word embedding model to be used during training is also treated as a hyperparameter.

| Hyperparameter | Range |
|---|---|
| adam_b2 | 0.9,0.999 |
| n_dense_output | 100,200,300,400 |
| keep_prob (dropout) | 0.4,0.5,0.6,0.7,0.8,0.9 |
| batch_size | 50,100,150 |
| learning_rate | 0.0001,0.001 |
| word_embedding | godin, shin |
| n_filters | 100,200,300,400 |
| filter_sizes | [1,2,3,4,5], [2,3,4,5,6], [3,4,5,6,7], [1,2,2,2,3], [2,3,3,3,4], [3,4,4,4,5], [4,5,5,5,6] |

**Table 2.** Hyperparameter ranges used for random search permutations.

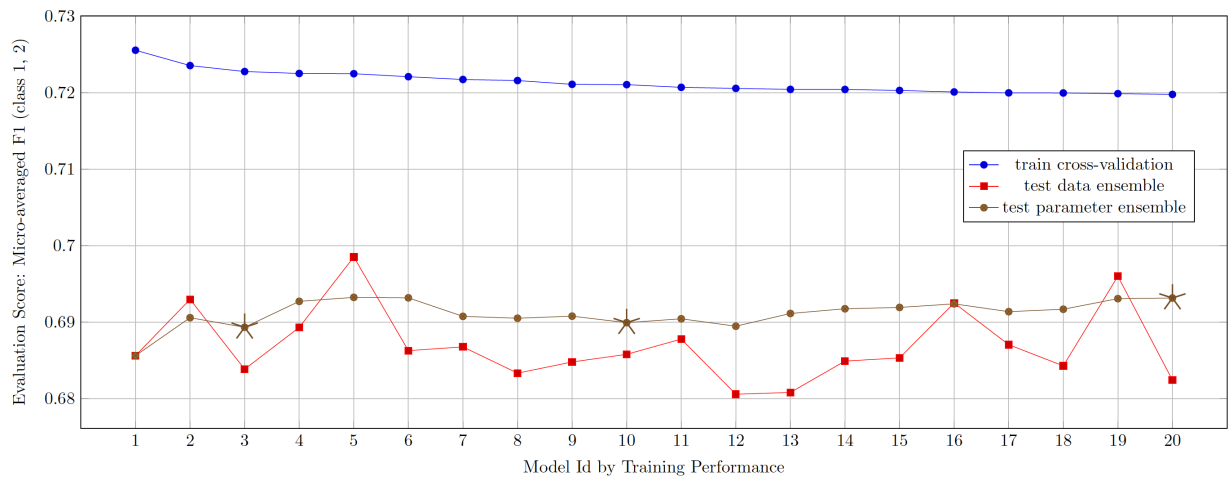

**Figure 2.** Individual 5-fold data ensemble and collective parameter ensemble (stacked ensemble) results for top 20 random search models. Models are sorted from left to right by decreasing 5-fold cross validation results.

**Results**

An ensemble of five CNNs is trained during 5-fold cross-validation training performed on our combined training dataset along with random search on the hyperparameter ranges. We train 99 such ensembles. The performance of the top 20 ensemble on the training data (blue) and on the test data (red) is shown in Figure 2. The models are arranged in the order of their decreasing training performance. We create stacked ensembles from these ensembles by taking top K ensemble models. We show the performances for such top K stacked ensembles (brown), as well. The detailed performances on the evaluation metrics of Top 3, Top 10 and Top 20 stacked ensembles are shown in

---
[4] https://spacy.io/
[5] https://www.tensorflow.org

Table 3, and denoted by stars in Figure 2. The stacked ensemble formed using top 20 best performing ensembles was submitted to the task, which achieved the best micro averaged F1 score on the task's test dataset. It can be also observed from Figure 2, that the fifth best ensemble model achieves the best scores on the test dataset. This proves an overall effectiveness of ensemble models in boosting performance on the present classification task.

| | Recall | | | Precision | | | F1 | | | $Recall_m$ | $Precision_m$ | $F1_m$ |
|---|---|---|---|---|---|---|---|---|---|---|---|---|
| | 1 | 2 | 3 | 1 | 2 | 3 | 1 | 2 | 3 | | | |
| Top 3 | 0.696 | 0.644 | 0.842 | 0.704 | 0.725 | 0.763 | 0.700 | 0.682 | 0.800 | 0.664 | 0.716 | 0.689 |
| Top 10 | 0.685 | 0.646 | 0.849 | 0.709 | 0.729 | 0.758 | 0.697 | 0.685 | 0.801 | 0.661 | 0.721 | 0.690 |
| Top 20 | 0.690 | 0.648 | 0.853 | 0.712 | 0.733 | 0.761 | 0.701 | 0.688 | 0.804 | 0.664 | 0.725 | 0.693* |

**Table 3.** Evaluation of submitted ensembles on test data. The *m* stands for micro average recall over class 1 and 2. The * marks the state-of-the-art micro averaged F1 on the task's dataset achieved by our best model.

**Conclusion and Future Work**

By participating in this shared task we showed the generic effectiveness of CNNs and ensembles on identification of personal medication intake from Twitter posts. Our proposed architecture of stacked ensemble of shallow CNNs, out-performed other models submitted in the task. This provided an empirical evaluation of our initial aim of combining ensembles with CNNs along with training the models using random search on the hyperparameters. In the future, we plan to work more on hyperparameter tuning using random search and various other search procedures and analyze their effectiveness. Instead of using pre-trained word embeddings it would also be interesting to look at the performance of our models by training word and phrase embeddings on a domain specific dataset of tweets. We would also like to formalize the architecture of stacked ensembles of CNNs and compare our models with an exhaustive set of other deep learning as well as traditional machine learning models.